\documentclass{article}
\usepackage[sglblindworkshop, nonatbib, final]{neurips_2025}

\usepackage[numbers, compress]{natbib}

\usepackage[utf8]{inputenc} % allow utf-8 input
\usepackage[T1]{fontenc}    % use 8-bit T1 fonts

\usepackage{algorithm}
\usepackage{algorithmic}
\usepackage{subcaption}
\usepackage{hyperref}       % hyperlinks
\usepackage{url}            % simple URL typesetting
\usepackage{booktabs}       % professional-quality tables
\usepackage{amsfonts}       % blackboard math symbols
\usepackage{nicefrac}       % compact symbols for 1/2, etc.
\usepackage{microtype}      % microtypography
\usepackage{xcolor}         % colors
\usepackage{graphicx}
\usepackage{multirow}
\usepackage{booktabs}
\usepackage{amsmath}
\usepackage{amssymb}
\usepackage{mathtools}
\usepackage{amsthm}
\usepackage[capitalize,noabbrev]{cleveref}
\title{Particle Monte Carlo methods for Lattice Field Theory}

\workshoptitle{Frontiers in Probabilistic Inference: Sampling Meets Learning}

\author{David Yallup \\ % \thanks{Corresponding author}\\
Kavli Institute for Cosmology Cambridge, \\
University of Cambridge \\
\texttt{dy297@cam.ac.uk}
}

\begin{document}

\maketitle

\begin{abstract}
High-dimensional multimodal sampling problems from lattice field theory (LFT) have become important benchmarks for machine learning assisted sampling methods. We show that GPU-accelerated particle methods, Sequential Monte Carlo (SMC) and nested sampling, provide a strong classical baseline that matches or outperforms state-of-the-art neural samplers in sample quality and wall-clock time on standard scalar field theory benchmarks, while also estimating the partition function. Using only a single data-driven covariance for tuning, these methods achieve competitive performance without problem-specific structure, raising the bar for when learned proposals justify their training cost.
\end{abstract}

\section{Introduction}\label{sec:intro}

Computing observables in strongly coupled quantum field theories (QFTs) remains a central challenge in fundamental physics. When perturbation theory breaks down, lattice field theory (LFT) offers a non-perturbative formulation via discretized Euclidean path integrals and has long underpinned precision studies in particle physics~\cite{Gross:2022hyw,Kronfeld:2012uk}. In parallel, LFT-inspired statistical problems have recently become established benchmarks for machine-learning–based sampling methods~\cite{Albergo:2019eim,Hackett:2021idh,Abbott:2022hkm,Abbott:2022zsh,Albergo:2022qfi,holderrieth2025leaps}, and ML approaches are now increasingly deployed within the LFT community itself~\cite{Boyda:2022nmh,Kanwar:2024ujc,Bialas:2025ood}. These models provide attractive testbeds because they are high dimensional, induce strong parameter correlations, exhibit phase transitions, and are frequently multimodal—features known to stress both classical and modern samplers~\cite{he2025tricktreatpursuitschallenges,greniouxImprovingEvaluationSamplers2025b}.

Neural network constructed \emph{trivializing maps}~\cite{Luscher:2009eq,hoffman2019neutralizing,PhysRevD.104.094507} represent one successful line of attack within LFT, and connect directly to broader developments in \emph{neural samplers} inspired by advances in normalizing flows, diffusion models, and neural transport~\cite{diffusionSong2020,papamakarios2021normalizing,lipman2023flow,midgley2023flowannealedimportancesampling,matthews2023continualrepeatedannealedflow,Gabrie:2021tlu,blessing2025underdamped,richter2024improved,vargas2025transportmeetsvariationalinference,albergo2025netsnonequilibriumtransportsampler,grenioux2024stochasticlocalizationiterativeposterior,zhang2024diffusion}. These methods aim to amortize sampling by learning transformations of an unnormalized target distribution. While promising results have been obtained for a range of LFT benchmarks, open questions remain concerning their scalability and robustness. Frontier QCD calculations require sampling in $\sim10^8$ dimensions, far beyond current neural deployments. Auxiliary learning paradigms may help bridge this gap~\cite{Albandea:2023ais,Gerdes:2022eve}; however, robustly addressing topological challenges at increasingly fine spacing remains elusive. Moreover, many neural samplers rely on carefully engineered equivariance or problem-specific symmetry handling. How these approaches transfer to domains where the relevant symmetries are unknown \emph{a priori}, or indeed how well they are then actually learning a trivializing map, remains poorly understood.

These considerations motivate a re-examination of classical baselines under modern computational constraints. Although traditional Monte Carlo algorithms were designed for CPUs and often rely on control-flow–heavy updates that map poorly to accelerators~\cite{pmlr-v151-hoffman22a,pmlr-v130-hoffman21a,rioudurand2023adaptivetuningmetropolisadjusted}, recent evidence shows that careful GPU implementations of classical techniques, such as non-reversible parallel-tempering schemes~\cite{Syed_2021}, can outperform state-of-the-art neural samplers even on synthetic multimodal problems~\cite{he2025tricktreatpursuitschallenges}. Motivated by this, we consider a broad family of accelerator-friendly \emph{particle Monte Carlo} methods~\cite{doucet2001introduction,delmoral,skilling} and evaluate them on a canonical multimodal LFT benchmark: the broken-phase scalar $\phi^4$ theory~\cite{Hackett:2021idh}. Our findings show that simple particle-based methods—implemented in a batched, GPU-native framework and tuned only through a single data-driven covariance estimate—provide a surprisingly strong baseline. Particularly if we treat the problem as truly black-box. We introduce the $\phi^4$ benchmark and its sampling challenges in \cref{sec:challenges}, present our empirical evaluation in \cref{sec:experiments}, and discuss implications for future work in \cref{sec:discussion}.

\section{Sampling Challenges in Lattice Field Theory}
\label{sec:challenges}

This section introduces the benchmark, outlines its characteristic sampling pathologies, and summarizes the particle-based Monte Carlo baselines we evaluate. Our primary experiments use a $10{\times}10$ lattice (state dimension $V=L^2$) and we later consider larger $L$ to assess scaling behaviour.

\paragraph{Scalar $\phi^4$ theory in 1+1D.}
Following Hackett et al.~\cite{Hackett:2021idh}, we study the $\mathbb{Z}_2$–symmetric real scalar $\phi^4$ model in $1{+}1$ dimensions. After Wick rotation, we work on a two-dimensional Euclidean lattice $\Lambda=\{0,\ldots,L{-}1\}^D$ with $D=2$, lattice spacing $a=1$, and periodic boundary conditions $\phi_{x+L\hat\mu}=\phi_x$ for unit vectors $\hat\mu$. The Euclidean action is
\begin{equation}\label{eq:action}
S_E(\phi)
= \sum_{x\in\Lambda} \left[
\tfrac{1}{2} \sum_{\mu=1}^{D} (\phi_{x+\hat\mu}-\phi_x)^2 \;+\;
\tfrac{1}{2} m_0^2\,\phi_x^2 \;+\; \lambda\,\phi_x^4
\right],
\end{equation}
and defines the Boltzmann target
\[
p(\phi) = \frac{1}{Z}\,e^{-S_E(\phi)}, \qquad
Z = \int_{\mathbb{R}^V} e^{-S_E(\phi)}\,d\phi.
\]
We typically take $m_0^2=-4$ and $\lambda=1$, giving a double-well potential and a bimodal magnetization distribution. As an order parameter we use the lattice-averaged magnetization
\[
\langle \phi\rangle = \tfrac{1}{V}\sum_{x\in\Lambda}\phi_x ,
\]
whose marginal is bimodal due to the exact $\mathbb{Z}_2$ symmetry $\phi\mapsto -\phi$. Example field configurations at $\beta=0$ and $\beta=1$ appear in \cref{fig:field_configs}.

To incorporate annealing or tempering, we follow the standard construction introducing an inverse temperature $\beta\in[0,1]$ and a Gaussian reference
\[
U_0(\phi)=\tfrac{1}{2}\sum_x \phi_x^2,\qquad
p_\beta(\phi) \propto \exp\!\big(-[U_0(\phi)+\beta\,S_E(\phi)]\big).
\]
Thus $\beta=0$ yields the Gaussian base distribution and $\beta=1$ recovers the physical ensemble. In physics applications, $\beta$ is often used to interpolate between the free and interacting theories while retaining physical meaning—e.g.\ in scans toward criticality (see \cref{app:alt_param}). The auxiliary model retains the essential sampling challenges, and the severity of bimodality can be tuned through $(m_0^2,\lambda)$.

\paragraph{Sampling challenges}
This benchmark concentrates several difficulties typical of lattice field theory: (i) multimodality and rare transitions between symmetry-related minima, which hinder global mixing; (ii) strong nearest-neighbor couplings that induce long autocorrelation times for local updates and lead to critical slowing down~\cite{WOLFF199093}; and (iii) barriers that become more pronounced as $L$ increases. These properties stress both neural and classical samplers and motivate hardware-accelerated, particle-based baselines evaluated in this work. Related phenomena in gauge theories include topological freezing—suppressed tunneling between topological sectors at fine lattice spacing—which further complicates sampling~\cite{DelDebbio:2004xh}. As a natural extension, one can consider the Schwinger model (lattice QED in $1{+}1$D)~\cite{Albergo:2022qfi}, which introduces gauge constraints and fermionic degrees of freedom and further amplifies these challenges.

\paragraph{Particle-based Monte Carlo methods.}
Hamiltonian (Hybrid) Monte Carlo (HMC) remains the standard tool for large-scale lattice simulations~\cite{DUANE1987216}, yet near criticality it also suffers from long autocorrelation times and—in more realistic gauge theories—topological freezing. Motivated by the need for scalable, accelerator-compatible alternatives, we consider simple particle-based methods with minimal tuning. These methods combine sequential importance tempering with lightweight MCMC rejuvenation steps and parallelize naturally on GPUs.

We evaluate three sequential Monte Carlo (SMC) samplers~\cite{chopin2020introduction} and one nested sampling (NS) baseline, implemented via \texttt{blackjax}~\cite{cabezas2024blackjax}: (i) SMC–RW, an adaptive tempered SMC scheme~\cite{fearnhead2010adaptivesequentialmontecarlo} using Gaussian random-walk proposals scaled by the particle covariance; (ii) SMC–IRMH, using an independent Metropolis–Hastings proposal tuned to the particle covariance, similar to a basic annealed importance sampling scheme~\cite{neal1998annealedimportancesampling}; (iii) SMC–HMC, replacing random-walk rejuvenation with short HMC trajectories preconditioned via window adaptation~\cite{stan_reference_manual_hmc_parameters}; and (iv) NS, nested sampling using slice-sampling rejuvenation~\cite{nealslicesampling} with direction covariance tuned to the particle cloud~\cite{yallup2025nested}. Additional details appear in \cref{app:alg}.

\paragraph{Black-box samplers.}
A central theme of this work is the comparison between methods that exploit explicit problem structure and those that do not. All particle MC methods considered here are intentionally configured as \emph{black-box samplers}: they incorporate no symmetry-aware moves or physics-specific modifications. To obtain a high-fidelity reference, we include an Adaptive HMC (AHMC) control that \emph{does} exploit the exact $\mathbb{Z}_2$ symmetry via occasional global flips $\phi\mapsto -\phi$, which greatly improves mixing. Such symmetry moves could be added to particle methods as well, but are intentionally omitted to assess their intrinsic performance. For completeness, we compare against the neural sampler of~\cite{Gerdes:2022eve} (see \cref{app:neural}) and additionally include a purely black-box neural sampler without encoded symmetries, clarifying the gap between physics-informed and genuinely symmetry-agnostic learned proposals.

\section{Experiments}
\label{sec:experiments}

We first evaluate the sampling efficiency and quality of four particle-based methods alongside two variants of a neural sampler on the bimodal $\phi^4$ scalar field theory at lattice size $L\times L$ with $L=10$, mass parameter $m_0^2=-4.0$, and quartic coupling $\lambda=1.0$. Our goals are to assess both sample quality from the physical ensemble at $\beta=1$ and computational efficiency. As a high-fidelity reference, we run a long Adaptive HMC (AHMC) chain. All experiments are run on an NVIDIA L4 GPU. The summary table of results is given in \cref{tab:stage1_comprehensive}. Further detail is given in \cref{app:experiments}.

\begin{table*}[!ht]
\centering
\caption{Sampling Quality and Performance Comparison. We report the mean and standard deviation of quality metrics computed against 10 reference sample sets from a long AHMC chain. MMD and $W_2$ measure discrepancy from ground truth (lower is better). The AHMC row establishes a baseline for inherent variance between different sets of true samples. All runtimes are in seconds on an NVIDIA L4 GPU (*AHMC takes around 200s to run on this GPU, the time listed is on CPU).}
\label{tab:stage1_comprehensive}
\begin{tabular}{lcccc}
\toprule
Method & MMD$\times 1000$ & $W_2 \times 100$ & $\log Z$ & Runtime (s) \\
\midrule
AHMC (control) & 2.96 ± 0.29 & 418.45 ± 1.09 & — & 7.4* \\
CNF~\citep{Gerdes:2022eve} & 6.04 ± 0.44 & 418.13 ± 1.07 & — & 1028.5 \\
\midrule
\multicolumn{5}{c}{\emph{Black-box methods}} \\
NS & 3.70 ± 0.28 & 412.90 ± 0.71 & -65.20 & 36.85 \\
SMC-RW & 7.45 ± 0.60 & 421.76 ± 0.61 & -65.26 & \textbf{7.74} \\
SMC-HMC & \textbf{3.43 ± 0.42} & 416.80 ± 0.53 & -65.64 & 8.15 \\
SMC-IRMH & 9.55 ± 0.85 & 425.77 ± 0.38 & -66.17 & 34.95 \\
CNF MLP & 9.64 ± 0.31 & \textbf{411.77 ± 1.43} & — & 2450.5 \\
\bottomrule
\end{tabular}
\end{table*}

\begin{figure}[htb]
  \centering
  \includegraphics[width=0.6\columnwidth]{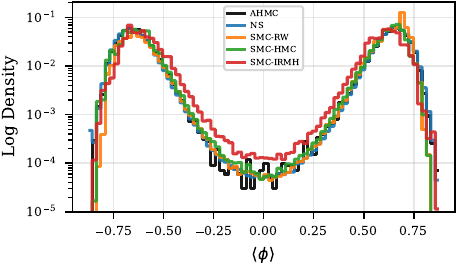}
  \caption{Posterior magnetization density histograms for AHMC (reference), NS, SMC-RW, SMC-HMC, and SMC-IRMH.}
  \label{fig:stage1_magnetization}
\end{figure}

We assess sample quality by comparing a number of standard quality metrics (see \cref{app:metrics}). We report means (and standard errors) over multiple batches to ensure stability. Summary statistics appear in Table~\ref{tab:stage1_comprehensive}, and magnetization density profiles for particle are shown in \cref{fig:stage1_magnetization}.

\paragraph{Additional experiments}
We conduct two supplementary studies. 
First, in \cref{app:scaling} we scale to larger lattices with $L\in\{15,18\}$, i.e., state dimension $V=L^2\in\{225,324\}$, finding that only SMC-IRMH starts to struggle, with all other algorithms scaling well and converging rapidly. 
Second, in \cref{app:asymptotic} we consider how the hyperparameters of the particle methods can be tuned to achieve asymptotic convergence, demonstrating that even gradient-free SMC-RW can be tuned to achieve high accuracy using simple settings.

\section{Discussion}\label{sec:discussion}
Our experiments show that simple, minimally tuned particle-based methods perform strongly on the bimodal $\phi^4$ benchmarks. As illustrated in \cref{fig:stage1_magnetization,fig:ess_magnetization_comparison,fig:magnetization_scaling}, all methods except SMC–IRMH at the largest scales produce high-quality samples in the broken phase, with close agreement to the AHMC reference. \cref{tab:stage1_comprehensive} indicates that the three local walk methods (SMC–RW, NS, SMC–HMC) offer complementary strengths and achieve accuracy consistent with the reference across our metrics, while \cref{tab:stage2_ess_analysis} shows that modest tuning of a few hyperparameters yields further gains with favorable runtime trade-offs, particularly when compared to the multiple GPU hours neural samplers take to train on similar problems~\cite{PhysRevD.104.094507}. Scaling results in \cref{tab:scaling_analysis_full} demonstrate competitive performance at larger lattice sizes. It is also possible to simply encode the $\mathbb{Z}_2$ symmetry into particle methods to improve mixing further.

The result of including a neural sampling baseline show two patterns. Firstly, that the cost of training these proposals vastly exceeds the cost of even un-tuned stochastic methods. Secondly the neural baselines fail to match the quality metrics of stochastic sampling despite the increased cost. The best $W_2$ result in \cref{tab:stage1_comprehensive} is achieved by a purely black-box neural sampler, but given the poor MMD and sampling efficiency this is likely due to over-concentration rather than a faithful representation of the target. After training, the neural proposal can be used in an MH independence sampler, giving rise to efficiencies of 85\% and less than 1\% for the CNF and CNF-MLP models respectively. While neural proposals can learn efficient representations of the target, and if symmetries are apriori known these can be encoded into the architecture to great effect, our results suggest two caveats. Firstly, apriori symmetries can be encoded into existing single particle HMC samplers just as easily in this case, solving the numerical sampling problem far more efficiently and scalably. Secondly, if these symmetries are not known, and we have to use black-box sampling (as is the case on many inverse problems), then simple particle methods can outperform neural proposals both in quality and wall-clock time without any problem-specific tuning.

\section{Conclusion}
Simple, minimally tuned particle-based samplers—using a data-driven particle covariance for scaling and preconditioning—achieve high accuracy and competitive or superior wall-clock performance on benchmark multimodal lattice problems. Across a number of metrics we demonstrate that local MCMC walks within SMC/NS can give state-of-the-art performance. These results challenge the view that particle methods are hard to tune or inherently hard to scale, and they raise the bar for when learned proposals justify their training cost on these tasks. Looking forward, we propose that initializing neural samplers with particle-based methods could be a promising direction.

% \section*{Software and Data}

% The core algorithm implementation has been prepared for submission to a widely used open source library. On publication, explicit experiment example code will similarly be made available. All data is composed of standard benchmark problems, which are open source and online.
    
    % If a paper is accepted, we strongly encourage the publication of software and data with the
    % camera-ready version of the paper whenever appropriate. This can be
    % done by including a URL in the camera-ready copy. However, \textbf{do not}
    % include URLs that reveal your institution or identity in your
    % submission for review. Instead, provide an anonymous URL or upload
    % the material as ``Supplementary Material'' into the OpenReview reviewing
    % system. Note that reviewers are not required to look at this material
    % when writing their review.
    
\section*{Acknowledgements}
We thank the authors of~\citep{Gerdes:2022eve} for making their neural sampler code publicly available. The authors were supported by the research environment and infrastructure of the Handley Lab at the University of Cambridge.
    
    % \textbf{Do not} include acknowledgements in the initial version of
    % the paper submitted for blind review.
    
    % If a paper is accepted, the final camera-ready version can (and
    % usually should) include acknowledgements.  Such acknowledgements
    % should be placed at the end of the section, in an unnumbered section
    % that does not count towards the paper page limit. Typically, this will
    % include thanks to reviewers who gave useful comments, to colleagues
    % who contributed to the ideas, and to funding agencies and corporate
    % sponsors that provided financial support.
    
    \section*{Impact Statement}
    
    This paper presents work whose goal is to advance the field of Machine Learning. There are many potential societal consequences of our work, none which we feel must be specifically highlighted here.

    % \bibliographystyle{unsrtnat}
    \bibliographystyle{icmlunsrt}
    
    \bibliography{references}
    % \bibliographystyle{../icml2025/icml2025}

    %%%%%%%%%%%%%%%%%%%%%%%%%%%%%%%%%%%%%%%%%%%%%%%%%%%%%%%%%%%%%%%%%%%%%%%%%%%%%%%
    %%%%%%%%%%%%%%%%%%%%%%%%%%%%%%%%%%%%%%%%%%%%%%%%%%%%%%%%%%%%%%%%%%%%%%%%%%%%%%%
    % APPENDIX
    %%%%%%%%%%%%%%%%%%%%%%%%%%%%%%%%%%%%%%%%%%%%%%%%%%%%%%%%%%%%%%%%%%%%%%%%%%%%%%%
    %%%%%%%%%%%%%%%%%%%%%%%%%%%%%%%%%%%%%%%%%%%%%%%%%%%%%%%%%%%%%%%%%%%%%%%%%%%%%%%
    \newpage
    \appendix
    \newpage

\section{Algorithm Details}\label{app:alg}

The outer kernels of all four main algorithms are largely similar. All feature the following free parameters: (i) population size, (ii) number of MCMC steps per tempering transition, and (iii) target effective sample size (ESS) for adaptive tempering (NS differs in using a fraction of particles to delete). These parameters trade off computational cost against accuracy: increasing the population size increases particle diversity but increases computational cost linearly; increasing the number of MCMC steps improves mixing with a linear cost increase; increasing the target ESS induces more tempering steps, improving accuracy at additional cost (for NS, decreasing the delete fraction toward zero has a similar effect).

We consider fairly standard settings for these parameters across our experiments.
\begin{itemize}
    \item Population size: $N=5000$ particles.
    \item MCMC steps per tempering transition: $n_\text{MCMC}=3 \times V$.
    \item Target ESS: $\text{ESS}_\text{target}=0.9$.
    \item NS delete fraction: $n_\text{delete}=0.5$.
\end{itemize}

In \cref{app:neural} we decrease $N$ to $1000$ to match the batch size of the forward ODE solver. In \cref{app:asymptotic} we systematically increase the target ESS to demonstrate asymptotic convergence and also set a higher baseline of $n_\text{MCMC} = 5 \times V$. The choice of inner kernel offers substantial opportunity for tuning; we forgo extensive tuning by basing all inner kernels on simple Gaussian covariances estimated from the particle population at each outer step, with the RW proposal scaled by $2.38^2/V$ to target optimal acceptance rates for RW MH algorithms~\cite{10.1214/ss/1177011137}. The HMC proposal uses a diagonal mass matrix tuned in warm-up and then frozen throughout the run. The HMC is run with a fixed trajectory of $20$ leapfrog steps, and the number of MCMC steps taken is reduced to a factor proportional to $V$, reflecting the higher per-step cost and faster mixing of each HMC step. Further gains could be made by online tuning of the mass matrix~\cite{buchholz2020adaptivetuninghamiltonianmonte}.

NS is distinctive, in that the NSS construction of~\cite{yallup2025nested} requires a successful MCMC step to satisfy the current likelihood constraint. Although the slice-sampling kernel (also tuned using the particle covariance) is configured to deliver efficiency comparable to the RW kernel, the additional work associated with enforcing the constraint (e.g., bracketing and shrinkage) typically leads to higher computational cost in practice.

The AHMC reference is comprised of a single long chain, with 5000 samples drawn in warm-up that is used to tune the diagonal of the mass matrix and step size, targeting an acceptance rate of 0.7 in the symplectic integrator. In particular we use the No-U-turn Sampling (NUTS) variant of~\cite{JMLR:v15:hoffman14a, 2017arXiv170102434B} HMC, leveraging the \texttt{blackjax} implementation of the window adaptation algorithm from Stan for tuning the HMC kernel parameters~\cite{stan_reference_manual_hmc_parameters}. The NUTS chain is run for $10^6$ iterations, thinning every $100$ steps to reduce autocorrelation. To ensure good mixing, we interleave standard NUTS trajectories with global $\mathbb{Z}_2$ flips $\phi\mapsto -\phi$ every $20$ steps. This is a simple problem-specific augmentation that is not available to the other methods, but it ensures that the AHMC reference is of high-fidelity. The resulting AHMC samples are then split into 10 separate tranches of $10^3$ samples, and used to compute error bars on quality metrics as in \cref{app:metrics}. All methods are run on an NVIDIA L4 GPU, except an M1 Studio Max that is used to compute CPU runtimes where noted. All experiments are run in \texttt{float32} precision.

\section{Neural Network training and Sampling Methods}\label{app:neural}

To test the performance of Neural Sampling methods, we employ the methods as outlined in~\citep{Gerdes:2022eve}. This trails a continuous normalizing flow (CNF) trained via minimization of the reverse Kullback-Leibler (KL) divergence. We perform the training on an NVIDIA L4 GPU, using a batch size of 1,024 samples and training for 1,000 steps. The flow is simulated using a simple Euler solver with a fixed step size of 0.02. We run on the same $10 \times 10$ $\phi^4$ lattice field theory with parameters $M^2=-4.0, \lambda=1.0$ as in the main text, and report the training time of the algorithm as by far the dominant factor. Such a flow can be used as a proposal to generate samples via IRMH, where the efficiency of the proposal is determined by the quality of the trained flow. Some of the sampling cost can be considered \emph{amortized} over multiple draws from the trained flow. To create a realistic \emph{black-box} version, we alter the neural network used to parameterize the flow to be a simple fully connected Multi-Layer Perceptron (MLP) with three hidden layers of 128 units each, using SiLU activations. We also use as input for this the raw field values as opposed to a physics-informed feature set as in~\citep{Gerdes:2022eve}. Both methods are found to have reasonably converged after this training time, through tracking the KL divergence and ESS of samples drawn from the trained flow.

\section{Additional Experimental Details}\label{app:experiments}

\begin{figure}[htb]
  \centering
  \includegraphics[width=0.9\columnwidth]{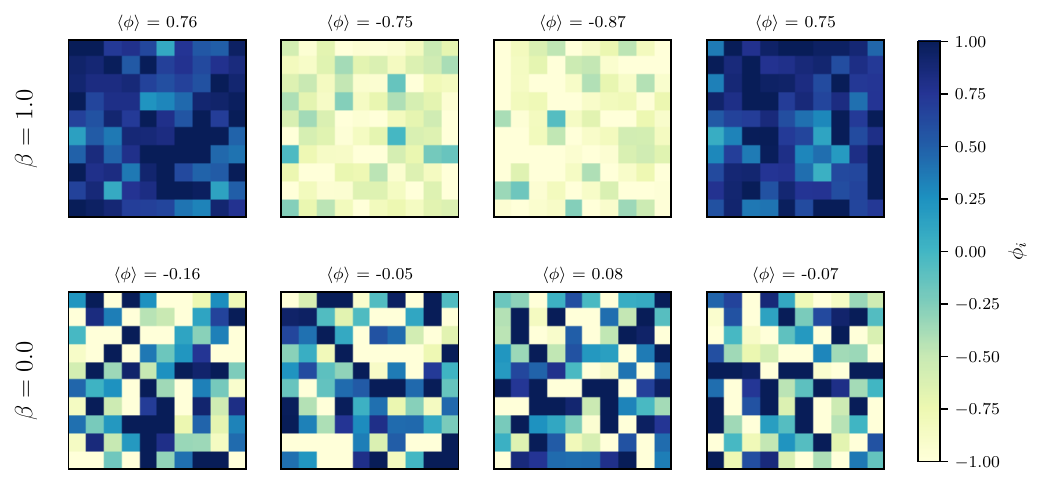}
  \caption{Eight $10\times 10$ field configurations sampled by NS at $\beta=1.0$ (top) and $\beta=0.0$ (bottom). Colors indicate site-wise fields $\phi_i$ with a shared colorbar; each panel is annotated with its mean magnetization $\langle\phi\rangle$.}
  \label{fig:field_configs}
\end{figure}

\begin{figure}[htb]
  \centering
  \begin{subfigure}{0.49\linewidth}
    \centering
    \includegraphics[width=\linewidth]{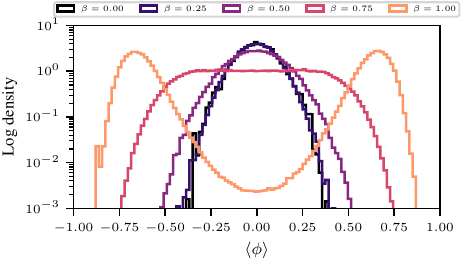}
    \caption{}
    \label{subfiga}
  \end{subfigure}
  \hfill
  \begin{subfigure}{0.49\linewidth}
    \centering
    \includegraphics[width=\linewidth]{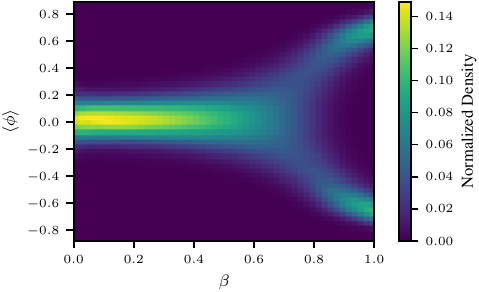}
    \caption{}
    \label{subfigb}
  \end{subfigure}
  \caption{Magnetization across inverse temperature $\beta$: (a) tempered histograms of $\langle\phi\rangle$ at five selected $\beta$ values; (b) density over $(\beta,\langle\phi\rangle)$ across the full temperature range.}
  \label{fig:phase}
\end{figure}

This section compiles additional quantitative and qualitative results referenced in the main text.
Table~\ref{tab:stage1_comprehensive} compares sampling quality and runtime.
Figure~\ref{fig:stage1_magnetization} shows posterior magnetization densities.
Figure~\ref{fig:phase} summarizes magnetization across inverse temperature, and Figure~\ref{fig:field_configs} visualizes representative field configurations.


Aside from features highlighted in the main text, we also point to some other interesting features of the results. Firstly, in introducing an auxiliary variable, as shown in \cref{fig:phase}, all methods are able to build a bridge that can connect the two modes of the distribution at $\beta=1.0$. This is a known feature of tempering methods, where we note that NS is not strictly a tempering method, but it does introduce an auxiliary variable in the form of the likelihood constraint, and is a key reason for their success on this problem. Secondly, we note that all methods also give much higher statistics in the low density region at $\beta=1.0$ than even the AHMC reference, this is again a feature of auxiliary variable methods that bridge from reference to target. Lastly, there are some statistical artifacts (bins with anomalously high weights) throughout the results. We consider that this is likely a fixable numerical artifact.

Beyond the main text, two patterns are noteworthy.
Methods that employ auxiliary variables—temperature schedules in the SMC variants and likelihood constraints in NS—induce sequences of intermediate distributions that effectively bridge the two magnetization modes at $\beta=1.0$ (see \cref{fig:phase}).
These methods also show greater coverage of the low-probability valley between modes at $\beta=1.0$ than the short AHMC reference, consistent with improved inter-mode mixing along the bridge.
Finally, there are some visible ``statistical artifacts'' (bins with anomalously high weights) throughout the results. We consider that this is likely a fixable numerical artifact from \texttt{float32} numerical precision of the weight calculations.

\begin{figure}[ht!]
    \centering
    \includegraphics[width=0.7\columnwidth]{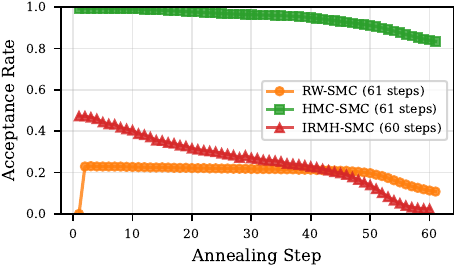}
    \caption{MCMC kernel acceptance rate during SMC annealing. This diagnostic plot shows the average acceptance rate of the internal MCMC kernel at each step of the annealing process for SMC-RW, SMC-IRMH, and SMC-HMC. This metric reveals the efficiency of the proposal mechanism as the algorithm transitions from the prior to the posterior.}
    \label{fig:acceptance_rate_diagnostic}
\end{figure}

We also include a diagnostic plot of the MCMC kernel acceptance rate during SMC annealing in \cref{fig:acceptance_rate_diagnostic}. This plot shows the average acceptance rate of the internal MCMC kernel at each step of the annealing process for SMC-RW, SMC-IRMH, and SMC-HMC. This metric reveals the efficiency of the proposal mechanism as the algorithm transitions from the prior to the posterior. We see that SMC-IRMH starts with a high acceptance rate but quickly drops off as the annealing progresses, indicating that the independent proposal becomes less effective in exploring the target distribution. In contrast, SMC-RW and SMC-HMC maintain more stable acceptance rates throughout the annealing process, suggesting that their local proposal mechanisms are better suited to adapt to the changing landscape of the target distribution.

\section{Scaling with Lattice Size}\label{app:scaling}

\begin{figure*}[ht]
    \centering
    % --- Panel (a) ---
    \begin{subfigure}{0.6\textwidth}
        \includegraphics[width=\linewidth]{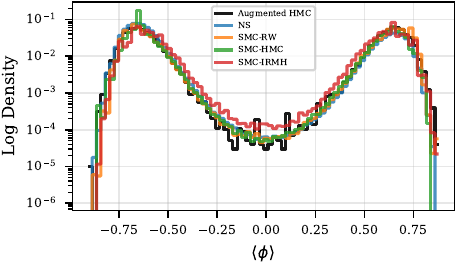}
        \caption{$10 \times 10$ Lattice}
        \label{fig:mag_L10}
    \end{subfigure}
    \hfill % Adds horizontal space
    % --- Panel (b) ---
    \begin{subfigure}{0.6\textwidth}
        \includegraphics[width=\linewidth]{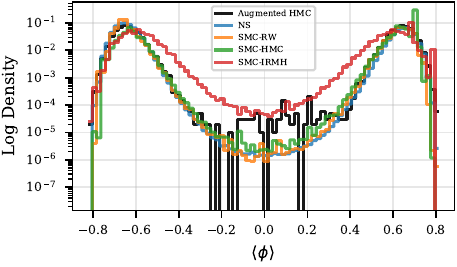}
        \caption{$15 \times 15$ Lattice}
        \label{fig:mag_L15}
    \end{subfigure}
    \hfill % Adds horizontal space
    % --- Panel (c) ---
    \begin{subfigure}{0.6\textwidth}
        \includegraphics[width=\linewidth]{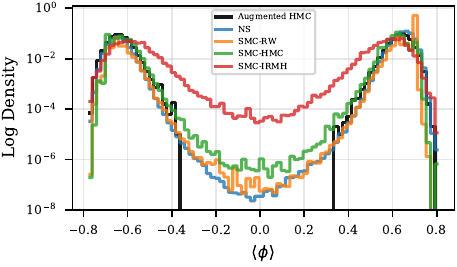}
        \caption{$18 \times 18$ Lattice}
        \label{fig:mag_L18}
    \end{subfigure}
    \caption{Posterior magnetization density as a function of increasing lattice size. Each panel compares the ability of the sampling methods to capture the bimodal posterior distribution. As the lattice dimension grows, the two modes become more sharply peaked and separated, increasing the difficulty of the sampling problem. Quantitative details relating to this scaling are listed in \cref{tab:scaling_analysis_full}.}
    \label{fig:magnetization_scaling}

\end{figure*}

We increase the lattice size from $L=10$ to $L\in\{15,18\}$, corresponding to state dimensions $V=L^2=100, 225,$ and $324$.
We fix the physical parameters at $m_0^2=-4.0$ and $\lambda=1.0$; as $L$ grows, the posterior magnetization distribution becomes more sharply bimodal, increasing sampling difficulty.
All algorithmic parameters are held fixed (population size $N=5000$, MCMC steps per tempering transition $n_\text{MCMC}=3\times V$, target ESS $\text{ESS}_\text{target}=0.9$).
Results are summarized in \cref{fig:magnetization_scaling} and \cref{tab:scaling_analysis_full}.
SMC-IRMH degrades at the larger sizes, whereas NS, SMC-RW, and SMC-HMC maintain good performance and converge rapidly (as reflected by the mode-coverage metrics).

For these runs we report mode-coverage metrics in place of full-distribution discrepancies because storing all field configurations becomes costly at larger $L$; comprehensive evaluation with full state storage is left to future work.
A higher-fidelity scaling study is needed to characterize computational scaling precisely, but runtimes remain modest across the tested sizes (see \cref{tab:scaling_analysis_full}).
At $L=18$, the AHMC (reference) chain with $10^6$ draws does not populate the low-density valley between modes (\cref{fig:mag_L18}), indicating how sharply separated this problem has become. In theory all these methods can be scaled further, but more extensive tuning may be required to maintain high accuracy.

% Prefer \input to avoid page breaks from \include
\begin{table*}[htb]
\centering
\caption{Scaling performance and quality across increasing lattice dimensions. We report runtime, two metrics for evaluating the exploration of the bimodal posterior (Mode Balance and Mode Ratio), and the estimated log evidence ($\log Z$). A mode balance of 1.0 indicates an equal number of samples were drawn from each of the two posterior modes. AHMC serves as the reference.}
\label{tab:scaling_analysis_full}
\vskip 0.15in
\begin{tabular}{clrrrrr}
\toprule
Lattice Size & Method & Runtime (s) & Mode Balance & Mode Ratio & $\log Z$ \\
\midrule
\multirow{5}{*}{$10 \times 10$} & AHMC (control) & 216.7 & 1.001 & 1.000 & --- \\
 & SMC-RW & 6.7 & 0.986 & 1.096 & -65.2 \\
 & SMC-HMC & 6.7 & 0.999 & 1.764 & -65.7 \\
 & NS & 34.3 & 0.785 & 0.959 & -65.4 \\
 & SMC-IRMH & 10.7 & 0.998 & 1.220 & -66.0 \\
\midrule
\multirow{5}{*}{$15 \times 15$} & AHMC (control) & 237.6 & 1.000 & 1.000 & --- \\
 & SMC-RW & 21.5 & 1.001 & 1.452 & -146.4 \\
 & SMC-HMC & 25.5 & 0.998 & 1.823 & -148.5 \\
 & NS & 175.6 & 0.848 & 1.052 & -147.9 \\
 & SMC-IRMH & 47.5 & 0.999 & 0.957 & -151.7 \\
\midrule
\multirow{5}{*}{$18 \times 18$} & AHMC (control) & 237.4 & 1.000 & 1.000 & --- \\
 & SMC-RW & 46.0 & 1.007 & 2.063 & -209.9 \\
 & SMC-HMC & 73.6 & 0.999 & 1.083 & -214.1 \\
 & NS & 414.2 & 1.189 & 1.016 & -212.9 \\
 & SMC-IRMH & 114.8 & 1.011 & 0.604 & -218.3 \\
\bottomrule
\end{tabular}
\end{table*}

\section{Asymptotic Convergence of Particle Monte Carlo Methods}\label{app:asymptotic}

To study asymptotic behavior, we revisit the weakest-performing method (in terms of quality, but fastest in runtime) from the main experiments and show that increasing the number of tempering steps (via a higher target ESS) yields near-convergent results while retaining substantially lower runtime than neural samplers reported in the main text.
We benchmark Sequential Monte Carlo with random-walk kernels (SMC-RW) against the same AHMC reference, running on the base $L=10$ lattice with the same setting for parameters $\lambda=1$, $m_0^2=-4$.

We evaluate three SMC-RW configurations with $N=5{,}000$ particles and an adaptive random-walk MCMC kernel using 500 inner transitions per tempering step, distinguished only by their target ESS for adaptive tempering:
\begin{enumerate}
  \item Standard: $\text{ESS}_\text{target}=0.90$
  \item High: $\text{ESS}_\text{target}=0.99$
  \item Max: $\text{ESS}_\text{target}=0.999$
\end{enumerate}
We note that there is a maximum target ESS of $1 - 1/N = 0.9998$ that can be achieved with $N$ particles, so the Max configuration is close to this limit.

\begin{figure}[ht!]
  \centering
  \includegraphics[width=0.7\columnwidth]{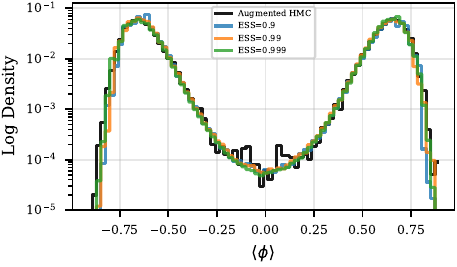}
  \caption{Effect of target ESS on posterior magnetization density for SMC-RW. Increasing $\text{ESS}_\text{target}$ from $0.90$ to $0.999$ induces a closer match to the bimodal distribution given by the AHMC (reference).}
  \label{fig:ess_magnetization_comparison}
\end{figure}

The results are summarized in \cref{tab:stage2_ess_analysis} and \cref{fig:ess_magnetization_comparison}.
As $\text{ESS}_\text{target}$ increases, the number of action (log-density) evaluations and hence runtime grow, but sample quality improves and approaches statistical compatibility with the AHMC control according to the reported metrics.

\begin{table*}[ht]
\centering
\caption{Table showing the trade-off between sample quality and computational cost for the SMC-RW sampler when varying the target Effective Sample Size (ESS). A higher ESS value forces more frequent resampling, which improves quality at the cost of longer runtimes.}
\label{tab:stage2_ess_analysis}
\vskip 0.15in
\begin{tabular}{lcccc}
\toprule
Method & MMD$\times 1000$ & Mode Balance & $\log Z$ & Runtime (s) \\
\midrule
Augmented HMC (Control) & 2.76 ± 0.28 & 1.000 & -- & 113.2 \\
\midrule
SMC-RW (ESS=0.900) & 7.18 ± 0.79 & 0.925 & -65.50 & 8.8 \\
SMC-RW (ESS=0.990) & 4.86 ± 0.23 & 0.957 & -65.48 & 18.4 \\
SMC-RW (ESS=0.999) & 3.28 ± 0.38 & 1.044 & -65.43 & 49.9 \\
\bottomrule
\end{tabular}
\end{table*}

\section{Sample Quality Metrics}
\label{app:metrics}

To provide a principled comparison of posterior sample quality, we employ two integral probability metrics to measure the statistical distance between the samples generated by a given method, $P$, and our reference distribution from AHMC, $Q$. For our analysis, we draw $n=m=1000$ samples for each comparison.

\subsection{Maximum Mean Discrepancy (MMD)}
We use the kernel two-sample discrepancy MMD~\cite{JMLR:v13:gretton12a}, which measures the distance between the mean embeddings of two distributions in a reproducing kernel Hilbert space (RKHS). For a characteristic kernel $k$, the squared MMD is
\[
\mathrm{MMD}^2(P,Q)
=\mathbb{E}_{x,x'\sim P}\big[k(x,x')\big]
-2\,\mathbb{E}_{x\sim P,\,y\sim Q}\big[k(x,y)\big]
+\mathbb{E}_{y,y'\sim Q}\big[k(y,y')\big].
\]
With a characteristic kernel, $\mathrm{MMD}(P,Q)=0$ if and only if $P=Q$. We use a Gaussian (RBF) kernel
\[
k(x,y)=\exp\!\bigl(-\|x-y\|^2/(2\sigma^2)\bigr),
\]
with bandwidth $\sigma$ set by the median heuristic. MMD is particularly effective at detecting differences in the shape and moments of distributions.

\subsection{2-Wasserstein Distance ($W_2$)}

We compute a Sinkhorn-regularized 2-Wasserstein cost. Given samples $X=\{x_i\}_{i=1}^n$ and $Y=\{y_j\}_{j=1}^m$ with uniform weights
$a=\tfrac{1}{n}\mathbf{1}_n$ and $b=\tfrac{1}{m}\mathbf{1}_m$, and squared Euclidean ground cost
$C_{ij}=\|x_i-y_j\|^2$, the entropic-regularized optimal transport problem is
\[
\mathrm{OT}_\varepsilon(X,Y)
=\min_{T\ge 0}\;\sum_{i,j} T_{ij}\,C_{ij}
+\varepsilon\sum_{i,j} T_{ij}\bigl(\log T_{ij}-1\bigr)
\quad\text{s.t.}\quad
T\,\mathbf{1}_m=a,\;\; T^\top \mathbf{1}_n=b.
\]
We solve this with the Sinkhorn algorithm~\cite{cuturi2013sinkhorn}, as implemented in \texttt{ott-jax}, using $\varepsilon=10^{-3}$. We report
\[
D_\varepsilon(X,Y)\;=\;\sqrt{\mathrm{OT}_\varepsilon(X,Y)},
\]
i.e., the square root of the regularized objective value (``Sinkhorn cost''). This is not the exact $W_2$ distance for finite $\varepsilon$, but $\mathrm{OT}_\varepsilon(X,Y)\to W_2(X,Y)^2$ as $\varepsilon\downarrow 0$. This effectively measures the distance between the empirical distributions of $X$ and $Y$, capturing differences in their support and overall geometry, with lower values indicating closer distributions.

\subsection{Mode balance and mode-height ratio}

\paragraph{Mode balance.}
Given a set of magnetization samples, we define the mode balance as the ratio of the number of positive samples to the number of negative samples. Samples exactly equal to zero are ignored. If there are no negative samples the score is $+\infty$ (all mass on the positive side); if there are no positive samples the score is $0$ (all mass on the negative side). Values near $1$ indicate a balanced occupancy of the two modes.

\paragraph{Histogram mode-height ratio.}
To compare peak sharpness against a reference, we construct histograms for the method and the reference over a common support and then compare their maximal heights on the positive and negative sides. Concretely, we form $B=80$ equally spaced bins over the combined range of the method and reference samples, compute the method histogram either with provided sample weights (accumulated in the log-domain via log-sum-exp for numerical stability and then normalized) or as unweighted counts (normalized), and compute the reference histogram as unweighted counts (normalized). We split the bins at zero magnetization into a positive side and a negative side; if a side has no bins, we use a small floor ($10^{-10}$) for its maximum. For each side, we take the maximum (mode height) of the method histogram and divide by the corresponding maximum of the reference histogram, and we report the geometric mean of these two ratios. A value of $1$ indicates matched peak sharpness in both modes relative to the reference; values greater (smaller) than $1$ indicate sharper (flatter) modes on average.

\section{Alternative parameterizations}
\label{app:alt_param}

To avoid confusion with the annealing parameter $\beta$ used in the main text, in this subsection we denote by $\kappa$ a \emph{physical} nearest-neighbor coupling that scales only the kinetic term.

A common, physically meaningful parameterization of lattice $\phi^4$ writes the action as
\begin{equation}
\label{eq:phi4-kappa-form}
S_E(\phi;\kappa,\lambda)
= \frac{\kappa}{2}\sum_{x,\mu} \big(\phi_{x+\hat\mu}-\phi_x\big)^2
\;+\; \sum_x \Big[\phi_x^2 + \lambda\,(\phi_x^2-1)^2\Big],
\end{equation}
see, e.g., \cite{PhysRevD.104.094507}. Expanding the nearest-neighbor term and collecting coefficients gives
\begin{align}
\frac{\kappa}{2}\sum_{x,\mu} \big(\phi_{x+\hat\mu}-\phi_x\big)^2
&= -\,\kappa \sum_{x,\mu}\phi_x\,\phi_{x+\hat\mu}
\;+\; \kappa D \sum_x \phi_x^2, \\
\phi_x^2 + \lambda(\phi_x^2-1)^2
&= \lambda \phi_x^4 \;+\; (1-2\lambda)\phi_x^2 \;+\; \lambda,
\end{align}
so that, up to an additive constant $|\Lambda|\,\lambda$ that only shifts $\log Z$,
\begin{equation}
\label{eq:phi4-kappa-expanded}
S_E(\phi;\kappa,\lambda)
= -\,\kappa \sum_{x,\mu}\phi_x\,\phi_{x+\hat\mu}
\;+\; (\kappa D + 1 - 2\lambda)\sum_x \phi_x^2
\;+\; \lambda \sum_x \phi_x^4 \;+\; \text{const}.
\end{equation}
This makes explicit that $\kappa$ strengthens nearest-neighbor correlations and shifts the effective quadratic term by $\kappa D$. We do not use this parameterization in our experiments. Our experiments target the Gaussian-regularized surrogate $\tilde p(\phi)\propto \exp(-[U_0(\phi)+S_E(\phi)])$ and use the main-text $\beta$ purely as an annealing parameter to bridge between distributions. We include \eqref{eq:phi4-kappa-expanded} to illustrate a physics-meaningful $\kappa$ that could be exploited in future work to scan for criticality with particle methods.

\end{document}